\documentclass{article}

     \usepackage[preprint]{neurips_2021}

\usepackage[utf8]{inputenc} 
\usepackage[T1]{fontenc}    
\usepackage{hyperref}       
\usepackage{url}            
\usepackage{booktabs}       
\usepackage{amsfonts}       
\usepackage{nicefrac}       
\usepackage{microtype}      
\usepackage{amsmath}
\usepackage{graphicx}
\usepackage{subfig}

\def\log{\mbox{log}}
\def\logit{\mbox{logit}}
\def\Nor{\mbox{Normal}}
\def\Ber{\mbox{Bernoulli}}

\def\Dir{\mbox{Dirichlet}}
\def\Gam{\mbox{Gamma}}

\def\ED{\mbox{ED}}
\def\RR{\mbox{RR}}
\def\A{\mbox{\bf{A}}}
\def\D{\mbox{\bf{D}}}
\def\E{\mathcal{E}}
\def\HH{\mbox{\bf{H}}}
\def\I{\mbox{\bf{I}}}
\def\KL{\mbox{KL}}

\def\R{\mathbb{R}}

\def\V{\mathcal{V}}
\def\W{\mbox{\bf{W}}}
\def\X{\mbox{\bf{X}}}

\def\h{\boldsymbol{h}}
\def\s{\boldsymbol{s}}

\def\z{\boldsymbol{z}}
\def\P{\mbox{P}}

\def\gam{\boldsymbol{\gamma}}
\def\del{\boldsymbol{\delta}}

\def\mmu{\boldsymbol{\mu}}
\def\sig{\boldsymbol{\sigma}}

\def\xxi{\boldsymbol{\xi}}
\def\ppsi{\boldsymbol{\psi}}
\def\ppi{\boldsymbol{\pi}}

\title{A Deep Latent Space Model for Graph Representation Learning}

\author{
Hanxuan Yang\\
\texttt{yanghanxuan2020@ia.ac.cn} \\
\And
Qingchao Kong\\
\texttt{qingchao.kong@ia.ac.cn} \\
\And
Wenji Mao\\
\texttt{wenji.mao@ia.ac.cn} \\
}

\begin{document}

\maketitle

\begin{abstract}

Graph representation learning is a fundamental problem for modeling relational data and benefits a number of downstream applications. Traditional Bayesian-based graph models and recent deep learning based GNN either suffer from impracticability or lack interpretability, thus combined models for undirected graphs have been proposed to overcome the weaknesses. As a large portion of real-world graphs are directed graphs (of which undirected graphs are special cases), in this paper, we propose a Deep Latent Space Model (DLSM) for directed graphs to incorporate the traditional latent variable based generative model into deep learning frameworks. Our proposed model consists of a graph convolutional network (GCN) encoder and a stochastic decoder, which are layer-wise connected by a hierarchical variational auto-encoder architecture. By specifically modeling the degree heterogeneity using node random factors, our model possesses better interpretability in both community structure and degree heterogeneity. For fast inference, the stochastic gradient variational Bayes (SGVB) is adopted using a non-iterative recognition model, which is much more scalable than traditional MCMC-based methods. The experiments on real-world datasets show that the proposed model achieves the state-of-the-art performances on both link prediction and community detection tasks while learning interpretable node embeddings. The source code is available at \url{https://github.com/upperr/DLSM}.

\end{abstract}

\section{Introduction}
\label{headings}

Graph representation learning is a fundamental problem for graph analysis and benefits a number of downstream applications, such as link prediction, community detection and node classification. Traditionally, a plethora of Bayesian-based random graph models have been proposed for learning graph representations \cite{holland1983stochastic, hoff2002latent, karrer2011stochastic, sewell2015latent}. Despite the ideal theoretical properties of these methods, they are impractical to model large-scale networks due to the expensive iterative inference procedures. Taking advantage of the powerful representation learning ability of deep learning, the graph neural networks (GNN) have been proposed to learn the topology of graph-structured data \cite{kipf2016semi, hamilton2017inductive, wang2018graphgan, velivckovic2018graph}. However, these deep learning methods usually bring about the interpretability issues.

To take the advantages of both Bayesian-based and deep learning based graph models and foster their mutual facilitation, recently, some research attempts to combine the Bayesian methods with GNN, and designs deep latent variable based generative models for graph-structured data \cite{kipf2016variational, grover2019graphite, mehta2019stochastic, sarkar2020graph}. Although these combined models can preserve the effective representation ability and achieve better performances on downstream tasks, they are all designed for \textit{undirected} graphs. In contrast, a large portion of real-world graphs, such as retweet and citation networks, are \textit{directed} graphs, of which undirected graphs are special cases (i.e. an undirected edge can be seen as a two-way directed edge). The representation learning of directed graphs is particularly challenging since not only the existence but also the direction of edges need to be learned. Moreover, existing combined models for undirected graphs only consider the \textit{community structure} but ignore another important graph characteristic commonly observed in complex networks -- the \textit{degree heterogeneity}, which measures the skewness of the distributions of out- and in-degrees.

In this paper, we focus on directed graphs and better modeling various graph properties with nice interpretability, including community structure and degree heterogeneity. To this end, we propose the Deep Latent Space Model (DLSM) by marrying the graph convolutional networks (GCN) with the classic latent space approaches. Specifically, to generate an asymmetric adjacency matrix, the degree discrepancy of the two ends in each directed edge is utilized, which is measured by the pairwise node random factors as distance scales. Moreover, to accommodate an overlapping community structure, where nodes may belong to one or more communities, we use a binary latent vector to denote the membership of a node pertaining to each community. All latent variables are generated by a hierarchical variational auto-encoder (VAE) architecture, which consists of a GCN deterministic encoder and a stochastic decoder. For fast inference, the \textit{reparameterization trick} of each latent variable is leveraged to perform the stochastic gradient variational Bayes (SGVB) \cite{kingma2014efficient} algorithm, which is far more efficient and scalable compared with the traditional iterative inference methods.

The main contributions of this work are summarized as follows:
\begin{enumerate}
\item{We propose the DLSM to incorporate the classic Bayesian latent space approaches into deep learning frameworks and generate interpretable representations involving the community structure and degree heterogeneity of directed graphs.}
\item{A hierarchical VAE architecture is established to layer-wise connect a GCN encoder and a stochastic decoder, which enables the variational posteriors to depend on the approximate likelihood as well as priors from previous layers.}
\item{For fast inference, the SGVB algorithm using a recognition model is adopted as a more efficient and scalable alternative of the expensive MCMC inference.}
\item{Our experiments on five real world network datasets show that the learned representations are readily to be employed in downstream applications and achieve state-of-the-art performances in link prediction and community detection.}
\end{enumerate}

\section{Related Work}

In this section, we briefly review the traditional Bayesian-based random graph models and deep learning based generative graph representation methods.

\subsection{Bayesian-based random graph models}

Classic Bayesian-based random graph models have developed for decades and are still valued in modeling relational data. One of the most well-known methods is the stochastic blockmodel (SBM) \cite{holland1983stochastic}, which generates a latent variable indicating the community membership of each node. Following SBM, a large number of variants have been proposed, such as allowing an overlapping community structure \cite{miller2009nonparametric} and involving degree heterogeneity \cite{karrer2011stochastic, sengupta2018block}. Another stream of research is the latent space models (LSM) \cite{hoff2002latent}, which endows each node with a low-dimensional latent variable to represent the node's position in a network. The most significant difference between these two methods is that LSM use the node similarity to measure the divergences between nodes, rather than communities, and thus can easily adapt to the complex directed networks. More recently, \cite{handcock2007model} involves the community structure within LSM using mixture Normal priors of latent positions. Furthermore, \cite{krivitsky2009representing} and \cite{sewell2015latent} introduce a pair of random variables in each link to model the degree heterogeneity of nodes. Although these above methods provide good theoretical properties and learn interpretable node embeddings, they rely on either the MCMC posterior sampling or mean-field variational inference with dramatically high computational complexity, and thus are usually unfeasible when modeling large scale networks.

\subsection{Deep learning based graph models}

The intriguing achievements of deep learning models for learning representations of Euclidean data have encouraged efforts to employ them on graph-structured data. The earliest attempts include DeepWalk \cite{perozzi2014deepwalk} and node2vec \cite{grover2016node2vec}, both of which encode local relations as node representations by conducting random walks on a graph. More recently, many GNN-based deep generative models have been proposed. \cite{kipf2016semi} first leverages the spectral GCN to generate node embeddings using the global topology of a graph. GraphSAGE \cite{hamilton2017inductive} samples a fixed-size neighborhood of each node for the inductive representation learning, which allows unseen nodes to be excluded during training. Later, GraphGAN \cite{wang2018graphgan} unifies a generator and a discriminator to learn node embeddings by playing a minimax game. The graph attention networks \cite{velivckovic2018graph} introduce the attention mechanism to aggregate the information from neighbors.

To take advantage of the nice theoretical properties of Bayesian methods and the powerful representation ability of deep learning methods, some research work tries to combine these two approaches for graph representation learning. The variational graph auto-encoders (VGAE) \cite{kipf2016variational} first generates Normal node embeddings using the variational auto-encoder (VAE) \cite{kingma2013auto}, which is then modified by Graphite \cite{grover2019graphite} with a perception component based on GCN. \cite{mehta2019stochastic} combines the classic SBM with GCN to learn a sparse real-valued node embedding interpreted as the memberships and strengths of each node belonging to different communities. Furthermore, \cite{sarkar2020graph} builds a sparse ladder VAE \cite{sonderby2016ladder} architecture to discover the communities at multiple levels of granularities. All of the existing deep learning based graph models, to the best of our knowledge, are only designed for undirected graphs without considering graph characteristics, such as degree heterogeneity. In this paper, we design a novel hierarchical variational auto-encoder architecture which combines the LSM and GCN to learn directed graph representations involving the community structure and degree heterogeneity.

\section{Preliminaries}

This section provides a formal definition of the graph representation learning problem concerned in this paper, followed by a brief introduction of the VAE method leveraged in our model.

\subsection{Problem definition}

Consider a directed network containing $n$ nodes. The data to be modeled include an asymmetric adjacency matrix $\A=(a_{ij})\in\{0,1\}^{n\times n}$, where each binary element $a_{ij}$ denotes the presence ($1$) or not ($0$) of the directed edge from node $i$ to $j$, and a node attribute matrix $\X\in\R^{n\times p}$. Throughout this paper we assume all edges to be conditionally independent and satisfy $a_{ij}\vert\Theta\sim\Ber(p_{ij})$, where $\Theta$ is the collection of latent representations. The problem concerned is to learn the node representations (embeddings) which can best reconstruct the adjacency matrix.

\subsection{Variational auto-encoder}

The objective of VAE is to train a generative model (decoder) $p(\X\vert\Theta)$ which can generate $\X$ given latent variables $\Theta$. However, due to the complex non-linearity of neural networks, the true posteriors $p(\Theta\vert\X)$ are typically intractable and need to be approximated by the variational posteriors $q(\Theta\vert\X)$ using a recognition model (encoder).

\begin{figure}
	\centering
	\includegraphics[width=\columnwidth]{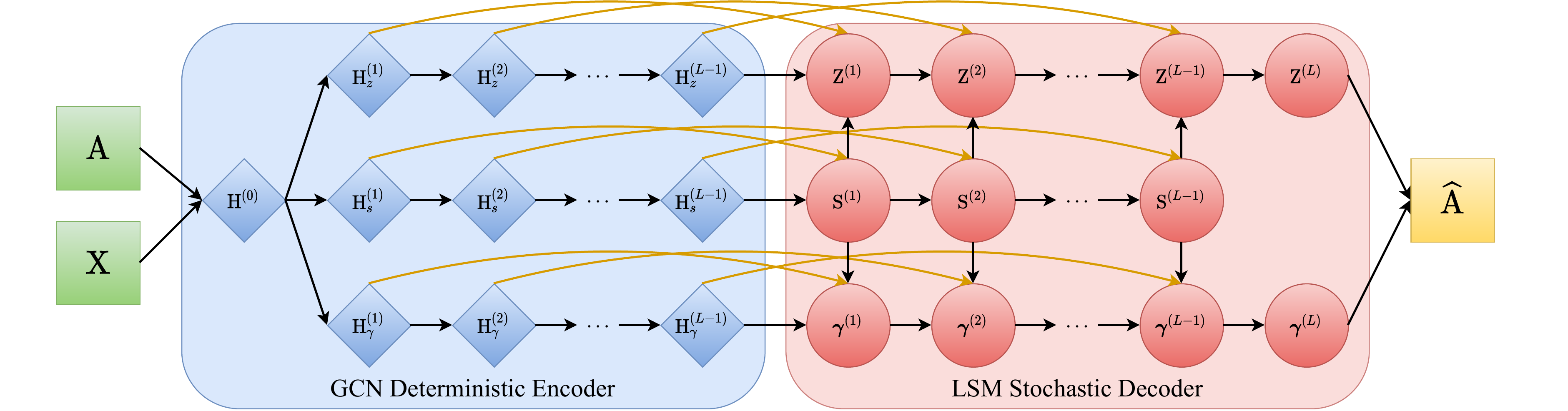}\par
	\caption{The architecture of hierarchical variational auto-encoder. The deterministic GNN (blue) encodes the adjacency matrix and attribute (if available) inputs as hidden states ${\mbox{\bf{H}}}^{(l)}$, which are then passed to the LSM decoder to generate the interpretable latent representations $\Theta^{(l)}$. For inference, the posterior variational distributions are also dependent on the GNN likelihood of the corresponding layers (orange arrows).}
	\label{hvae}
\end{figure}

One of the problems for the vanilla VAE is that the latent variables generated at deep layers tend to collapse into priors because of the noises accumulated during the multiple Monte Carlo sampling. Enlightened by the ladder VAE \cite{sonderby2016ladder}, which employs a bidirectional structure to layer-wise connect the encoder and decoder, here we propose a hierarchical architecture, as illustrated in Fig~\ref{hvae}. The encoder block takes the adjacency $\A$ and attribute matrices $\X$, if available, as inputs and learns a hidden state for each node using a directed GCN. Then, the decoder block recursively generates expressive latent variables from variational posteriors depending on not only the priors passed from previous layers of the decoder, but also the approximate likelihood learned by the encoder, which enables a deep hierarchy of the proposed model.

\subsection{Latent space model}

The proposed model learns interpretable node representations based on the classic LSM framework. Assume each node correspond to a latent position in an unobservable $D$-dimensional space, denoted as $\z_i\in\R^{D}$. The distances between latent positions indicate the relationships of nodes. The closer the latent positions are, the more possible the edge will exist. To accommodate the degree heterogeneity of directed networks, we propose a generalized form of LSM by involving two node-specific random factors, which serve as distance scales at each dimension of the latent space, i.e.
\begin{eqnarray}\label{lsm}
p_{ij}=\sigma(\beta_0-\beta_{out}\Vert\gam_i\odot(\z_i-\z_j)\Vert-\beta_{in}\Vert\del_j\odot(\z_i-\z_j)\Vert),
\end{eqnarray}
where $\odot$ denotes the element-wise multiplication and $\sigma(\cdot)$ is the sigmoid function. $\gam_i$ and $\del_i\in\R^{D}$ are the activity and popularity factors, the reverse of which represent the tendencies for a node to send and receive edges, respectively. The latent positions $\z_i$ are involved as Euclidean distances, indicating the relationships between the nodes. The global weights $\beta_{out}$ and $\beta_{in}$ are to measure the importance of out-degrees (activity) and in-degrees (popularity), and $\beta_0$ is a bias.


\paragraph{Special cases}
Though Eq.~(\ref{lsm}) is designed for directed networks due to the asymmetric structures of $p_{ij}$ and $p_{ji}$, it can be easily degraded to a symmetric structure by setting $\gam_i=\del_i$ and $\beta_{out}=\beta_{in}$ for undirected networks.

\section{Deep Latent Space Model}

We propose a hierarchical VAE architecture composed of a GCN encoder and an LSM decoder, as shown in Fig.~\ref{hvae}, to generate interpretable node representations via Monte Carlo sampling.

\subsection{LSM decoder}

We first introduce the hierarchical architecture of the stochastic decoder. Denoting $G_l$ as the size of the $l-$th decoder layer, there are four latent variables to generate, i.e. the latent positions $\z_i^{(l)}\in\R^{G_l}$, community memberships $\s_i^{(l)}\in\R^{G_l}$ and node random factors $\gam_i$, $\del_i^{(l)}\in\R^{G_l}$. All of these variables are randomly sampled from the variational distributions. At the first layer ($l=1$), the parameters of variational distributions are defined by priors solely, while for other layers ($l=2,\dots,L-1$), the parameters are obtained by the feedforward representations generated from previous layers as well as priors.

\paragraph{Latent positions}
The latent positions $\z_{i}^{(l)}$ measure the relationship between nodes through distances, generated as
\begin{eqnarray}\label{z_prior}
\z_{i}^{(l)}\sim \Nor(\s_i^{(l)}\odot f(\W_z^{(l-1)}\z_{i}^{(l-1)}),{\sig_{i}^2}^{(l)}),
\end{eqnarray}
where $f(\cdot)$ is a nonlinear activation function (e.g. leaky ReLU), ${\sig_{i}^2}^{(l)}\in\R^{G_l}$ is the prior variance to be specified, and $\W_z^{(l-1)}\in\R^{G_l\times G_{l-1}}$ is a weight matrix to transform the variable from dimension $G_{l-1}$ to $G_l$. For the convenience of notations, here the parameters of Normal distribution are symbolized as vectors, meaning the element-wise operations.

\paragraph{Community memberships}
The latent positions in Eq.~(\ref{z_prior}) are separated into different communities by the sparse binary variables $\s_i^{(l)}=(s_{i1},\dots,s_{iG_l})’$, thus an overlapping community structure can be adapted. Referring to \cite{miller2009nonparametric}, we employ the Indian buffet process (IBP) prior on $\s_i^{(l)}$ to learn the effective number of communities given a sufficiently large truncation parameter $G_l$, i.e.
\begin{eqnarray}\label{s_prior}
\pi_{ig}^{(l)}&=&\logit(\prod_{h=1}^gv_{ih}^{(l)}),\\
s_{ig}^{(l)}&\sim&\Ber(\sigma(\pi_{ig}^{(l)})).
\end{eqnarray}
Typically, the log odds $\pi_{ig}^{(l)}$ is generated using a stick-breaking construction, where $v_{ih}^{(l)}$ is drawn from a Beta distribution \cite{teh2007stick}. In our model we simplify such hierarchical prior structure by specifying a global $v$ for all nodes. At each layer, the community membership $s_{ig}^{(l)}$ denotes whether node $i$ belongs to community $g$, hence the size of each decoder layer $G_l$ can be explained as the number of communities. Additionally, the proposed HVAE architecture enables our model to detect community structures at multiple levels of granularities. Letting the layer sizes $G_l$ be downward increasing, the top layers indicate the coarse-grained communities and the bottom layers indicate the fine-grained communities.

\paragraph{Node random factors}
Considering the prevalent power-law of degrees, we propose the Dirichlet node random factors $\gam_i^{(l)}$, $\del_i^{(l)}\in\R^{G_l}$, so the sparse distributions of degrees can be modeled flexibly, i.e.
\begin{eqnarray}\label{gamma_prior}
\gam_i^{(l)}&\sim&\Dir(\xxi_i^{(l)}+\s_i^{(l)}\odot f(\W_{\gamma}^{(l-1)}\gam_i^{(l-1)})),\\
\del_i^{(l)}&\sim&\Dir(\ppsi_i^{(l)}+\s_i^{(l)}\odot f(\W_{\delta}^{(l-1)}\del_i^{(l-1)})),
\end{eqnarray}
where $\xxi_i^{(l)}$ and $\ppsi_i^{(l)}$ are prior parameters to be specified, and $\W_{\gamma}^{(l-1)}$, $\W_{\delta}^{(l-1)}\in\R^{G_l\times G_{l-1}}$ are weight matrices. Note that the social reach factors are dependent on the community membership $\s_i^{(l)}$ as well, which can be explained as only the random effects within the communities the node belongs to are involved in the latent representations.

At the output layer ($l=L$), the latent positions $\z_i^{(L)}$ and node random factors $\gam_i^{(L)}$, $\del_i^{(L)}$ are transformed from dimension $G_{L-1}$ to $D$ through a full connection layer, where weights are column-wise summing to one. Such transformation also changes the interpretation of the layer size, from the number of communities to the dimension of the latent space. Last, the adjacency matrix is reconstructed using Eq.~(\ref{lsm}).

\subsection{GCN encoder}

The proposed DLSM employs a deep encoder as a non-iterative recognition model to infer the parameters of posterior distributions. Assuming the mean-field approximation of the variational distributions, the true joint posterior of the latent variables $p_{\theta}(\Theta\vert\A,\X)$ can be approximated by a variational posterior $q_{\phi}(\Theta)$, where $\Theta={\{\z_i,\s_i,\gam_i,\del_i\}}_{i=1}^{n}$, $\theta$ and $\phi$ denote the generative (decoder) and inference (encoder) parameters to be trained, respectively. Then, the variational posterior is given as
\begin{align}\label{joint_post}
	q_{\phi}(\z,\s,\gam,\del)=&\prod_{i=1}^n\prod_{l=1}^Lq_{\phi}(\z_i^{(l)}\vert\h_i^{(l)},\z_i^{(l-1)})q_{\phi}(\s_i\vert\h_i^{(l)},\ppi_i^{(l-1)})\nonumber\\
	&q_{\phi}(\gam_i\vert\h_i^{(l)},\gam_i^{(l-1)})q_{\phi}(\del_i\vert\h_i^{(l)},\del_i^{(l-1)}),
\end{align}
where $\h_i^{(l)}\in\R^{K_l}$ is the output of the encoder and $K_l$ denotes the size of the $l$-th layer.

GCN has been proved effective in learning the topology of non-Euclidean data, and thus is an ideal choice for the encoder of our model. Referring to \cite{kipf2016semi}, we propose the directed GCN operator as
\begin{eqnarray}\label{gcn}
	\HH^{(l+1)}=f(\tilde{\D}_{out}\tilde{\A}\tilde{\D}_{in}\HH^{(l)}\W^{(l)}).
\end{eqnarray}
Here $\tilde{\A}=\A+\I_n$ and $\I_n$ is the $n$-dimensional identity matrix. $\tilde{\D}_{out}$ and $\tilde{\D}_{in}$ are diagonal matrices with elements as the out- and in-degrees of $\tilde{\A}$, respectively. $\HH^{(0)}=\X$ if $\X$ is available and $\HH^{(0)}=\I_n$ if not. During inference, the hidden states are passed to the corresponding layer of the decoder and then combined with the prior information from previous layers to generate the parameters of variational distributions. Note that the vanilla GCN used here can be substituted by any other GNN for directed networks.

\section{Inference}

We now introduce our fast inference method using the stochastic gradient variational Bayes (SGVB) algorithm \cite{kingma2014efficient}. Compared with the iterative methods such as MCMC adopted by traditional Bayesian random graph approaches, SGVB is much more efficient and scalable. Such method requires differential Monte Carlo expectations to perform backpropagation, thus the reparameterization trick for each of the latent variables is leveraged.

Letting $\z^{\ast}\in\R^{G_l}$ denote a vector with standard Normal elements, the latent positions are reparametrized as $\z_{i}^{(l)}=\hat{\mmu}_{i}^{(l)}(\z_{i}^{(l-1)},\s_i^{(l-1)},\h_i^{(l)})+\hat{\sig}_{i}^{(l)}(\z_{i}^{(l-1)},\s_i^{(l-1)},\h_i^{(l)})\odot\z^{\ast}$, where $\hat{\mmu}_{i}^{(l)}(\z_{i}^{(l-1)},\s_i^{(l-1)},\h_i^{(l)})$, $\hat{\sig}_{i}^{(l)}(\z_{i}^{(l-1)},\s_i^{(l-1)},\h_i^{(l)})$ are variational posterior parameters.

Following \cite{mehta2019stochastic}, the Bernoulli posterior $p_{\theta}(s_{ig}^{(l)})$ is approximated by the Binary Concrete distribution \cite{maddison2017concrete}, i.e.
\begin{eqnarray}\label{s_post}
\tilde{s}_{ig}^{(l)}=\frac{\hat{\pi}_{ig}^{(l)}(\s_i^{(l-1)},\h_i^{(l)})+\logit(u)}{\lambda},
\end{eqnarray}
where $u\sim U(0,1)$, $\hat{\pi}_{ig}^{(l)}(\s_i^{(l-1)},\h_i^{(l)})$ is a variational posterior parameter, $\lambda\in\R^+$ is the temperature to be specified, and then $s_{ig}^{(l)}=\sigma({\tilde{s}_{ig}^{(l)}})$.

Referring to the recent literature \cite{joo2020dirichlet}, we use multiple normalized Gamma variables with unified rate parameters to compose the Dirichlet distributions $p_{\theta}(\gam_i^{(l)})$ and $p_{\theta}(\del_i^{(l)})$, i.e.
\begin{eqnarray}\label{gamma_post}
\tilde{\gam}_i^{(l)}&\sim&\Gam(\hat{\xxi_i}^{(l)}(\gam_i^{(l-1)},\s_i^{(l-1)},\h_i^{(l)}),\boldsymbol{1}),\\
\tilde{\del}_i^{(l)}&\sim&\Gam(\hat{\ppsi_i}^{(l)}(\del_i^{(l-1)},\s_i^{(l-1)},\h_i^{(l)}),\boldsymbol{1}),
\end{eqnarray}
where $\hat{\xxi_i}^{(l)}(\gam_i^{(l-1)},\s_i^{(l-1)},\h_i^{(l)})$ and $\hat{\ppsi_i}^{(l)}(\del_i^{(l-1)},\s_i^{(l-1)},\h_i^{(l)})$ are variational posterior parameters. The node random factors are then derived from $\gam_i^{(l)}=\tilde{\gam}_i^{(l)}/\sum_j^n\tilde{\gam}_j^{(l)}$, $\del_i^{(l)}=\tilde{\del}_i^{(l)}/\sum_j^n\tilde{\del}_j^{(l)}$. In practice, the Dirichlet variables are magnified by $n$ times to avoid too small values of $\gam_i^{(l)}$ and $\del_i^{(l)}$ when $n$ is large.

In particular, the initial variational posterior parameters ($l=0$) are set as nonlinear combinations of $\h_i^{(1)}$ and $\h_i^{(L-1)}$, as illustrated in Fig.~\ref{hvae}.
 

The loss function is defined by minimizing the negative evidence lower bound (ELBO), i.e.
\begin{align}\label{loss}
\mathcal{L}=&\sum_{i=1}^n\sum_{l=1}^L(\KL[q_{\phi}(\z_i^{(l)})\Vert p_{\theta}(\z_i^{(l)})]+\KL[q_{\phi}(\s_i^{(l)})\Vert p_{\theta}(\s_i^{(l)})]
+\KL[q_{\phi}(\gam_i^{(l)})\Vert p_{\theta}(\gam_i^{(l)})]\nonumber\\
&+\KL[q_{\phi}(\del_i^{(l)})\Vert p_{\theta}(\del_i^{(l)})])-\sum_{i=1}^n\sum_{j=1}^n\mathbb{E}_q[\log p_{\theta}(a_{ij}\vert\Theta^{(L)}],
\end{align}
where $\KL[q(\cdot)||p(\cdot)]$ denotes the Kullback-Leibler (KL) divergence between $q(\cdot)$ and $p(\cdot)$. The second term is the cross entropy of adjacency matrix reconstruction. Here all of the true posteriors $p_{\theta}(\Theta^{(l)})$ and variational posteriors $q_{\phi}(\Theta^{(l)})$, except for the input layer ($l=1$), are conditioned on $\Theta^{(l-1)}$, which is omitted for simplification.

\section{Experiments}

To evaluate the performances of the proposed method, we conduct a series of experiments on real datasets. Two important downstream applications of graph analysis, i.e. link prediction and community detection, are considered in our experiments. Results show that the latent embeddings learned by DLSM can better represent directed graphs and significantly outperform baseline methods.

\subsection{Baselines}

We compare the proposed DLSM with a variant of our model by changing the Euclidean distance reconstruction layer to a inner product generator, i.e. $\P(a_{ij}=1\vert\Theta)=\sigma({(\beta_{out}\gam_i\odot\z_i)}^{\top}(\beta_{in}\del_j\odot\z_j))$ following \cite{mehta2019stochastic} and \cite{sarkar2020graph}, which we refer to as DLSM-IP. 


Other baselines include a traditional Bayesian random graph model, i.e. the popularity-scaled latent space model (PSLSM) \cite{sewell2015latent}, and four recent deep learning graph models, i.e. the variational auto-encoder on graphs (VGAE) \cite{kipf2016variational}, Graphite \cite{grover2019graphite}, the deep generative latent feature relational model (DGLFRM) \cite{mehta2019stochastic}, and the ladder Gamma variational auto-encoder for graphs (LGVG) \cite{sarkar2020graph}.


\subsection{Datasets}

The experiments of link prediction are conducted on five real world datasets. Specifically, Political blogs is a well-studied social network composed of U.S. political blog nodes and webpage links \cite{adamic2005political}. Kohonen is a citation network related to the self-organizing maps \cite{batagelj2006pajek}, where each node and directed edge represent a paper and a citation, respectively. CiaoDVD is a user–user trust network of an online DVD website in Britain \cite{guo2014etaf}. WikiVote is a network of Wikipedia users, where each directed edge represents a user voting on another to become the administrator \cite{leskovec2010predicting}. DBLP is a citation network within an authoritative database of computer science publications \cite{ley2002dblp}.

For community detection, we consider three datasets with ground-truth community labels. The nodes of political blogs have been labeled as ``liberal'' and ``conservative''. The Emails network consists of the members from 42 departments (communities) of a European research institution \cite{leskovec2007graph}. For this dataset, the communities with less than 30 nodes are excluded and finally 10 communities are retained. Lastly, the British MP is a network of politicians divided into 4 communities according to their parties \cite{greene2013producing}.

All of the networks have been preprocessed by omitting the isolated nodes and loops. In our experiments, they are randomly splitted as 85\% edges for training, 10\% edges for testing, and 5\% edges for validation. The descriptive statistics of the datasets are summarized in Table~\ref{datasets}.

\begin{table}
	\caption{Descriptive statistics of the real network datasets. $\vert\V\vert$ and $\vert\E\vert$ are the numbers of nodes and edges, respectively, CC is the clustering coefficient \cite{fagiolo2007clustering}, $d_{max}^{out}$ and $d_{max}^{in}$ are the maximal in-degree and out-degree of all nodes, respectively, $d_{avg}$ is the average degree of all nodes (average in-degree equals to average out-degree), $\ED=\sum_{i=1}^{n}\sum_{j=1}^{n}a_{ij}/(n(n-1))$ is the edge density, and $\RR=\sum_{i=1}^{n}\sum_{j=1}^{n}a_{ij}a_{ji}/\sum_{i=1}^{n}\sum_{j=1}^{n}a_{ij}$ is the reciprocal rate.}
	\label{datasets}
	\centering
	\begin{tabular}{lcccccccc}
		\toprule
		Dataset & $\vert\V\vert$ & $\vert\E\vert$ & CC & $d_{max}^{out}$ & $d_{max}^{in}$ & $d_{avg}$ & ED & RR\\
		\midrule
		Political blogs & 1,222 & 19,021 & 0.2459 & 256 & 337 &15.6 & 0.0127 & 0.2426\\
		Kohonen         & 3,772 & 12,731 & 0.1530 & 51  & 735 & 3.4 & 0.0013 & 0.0017\\
		CiaoDVD         & 4,658 & 40,133 & 0.1492 & 100 & 361 & 8.6 & 0.0019 & 0.3497\\
		WikiVote        & 7,115 &103,689 & 0.0896 & 893 & 457 &14.6 & 0.0020 & 0.0565\\
		DBLP            &12,590 & 49,744 & 0.0983 & 617 & 227 & 4.0 & 0.0003 & 0.0043\\
		Emails          &   986 & 24,929 & 0.4124 & 333 & 211 &25.3 & 0.0257 & 0.7112\\
		British MP      &   418 & 27,340 & 0.5314 & 201 & 303 &65.4 & 0.1569 & 0.5406\\
		\bottomrule
	\end{tabular}
\end{table}

\subsection{Link Prediction}

For link prediction, we use the area under the ROC curve (AUC) and average precision (AP) as evaluation metric. All baselines designed for undirected graphs are slightly modified by altering the original encoder to the proposed directed GCN given in Eq~\ref{gcn}. The experimental results of DLSM and the baselines are presented in Table~\ref{AUC} and Table~\ref{AP}, where the results are reported as the means and standard deviations of 10 independent random splits. Our model significantly outperforms the baselines on all datasets, especially the citation networks such as Kohonen and DBLP. The reason for this is that, these networks are almost unidirectional (i.e. the reciprocal rates are close to 0 as shown in Table~\ref{datasets}), since most publications can only be cited by later ones. The Bayesian PSLSM model is unpractical to fit large networks such as WikiVote and DBLP because of the high computational complexity of MCMC-based inference methods, while the SGVB method adopted by our model can handle large scale graphs efficiently. The experimental results also show that, the variant of our model, i.e. DLSM-IP, outperforms deep learning based methods on most datasets, demonstrating the importance of modeling degree heterogeneity of directed graphs.



\begin{table}
	\caption{AUC of link prediction.}
	\label{AUC}
	\centering
	\resizebox{\textwidth}{!}{
	\begin{tabular}{lccccc}
		\toprule
		         & Political blogs& Kohonen      & CiaoDVD    & WikiVote         & DBLP\\
		\midrule
		PSLSM    &$0.887\pm 0.005$&$0.903\pm 0.002$&$0.942\pm 0.002$& N/A              & N/A \\
		VGAE     &$0.848\pm 0.006$&$0.755\pm 0.008$&$0.846\pm 0.003$&$0.786\pm 0.002$&$0.707\pm 0.000$\\
		Graphite &$0.800\pm 0.011$&$0.819\pm 0.006$&$0.752\pm 0.003$&$0.735\pm 0.003$&$0.458\pm 0.019$\\
		DGLFRM   &$0.889\pm 0.006$&$0.783\pm 0.007$&$0.895\pm 0.002$&$0.887\pm 0.001$&$0.778\pm 0.000$\\
		LGVG     &$0.921\pm 0.005$&$0.855\pm 0.007$&$0.932\pm 0.002$&$0.943\pm 0.000$&$0.882\pm 0.000$\\
		\midrule
		DLSM-IP  &$0.894\pm 0.005$&$0.881\pm 0.007$&$0.964\pm 0.002$&$0.965\pm 0.001$&$0.920\pm 0.000$\\
		DLSM     &$\boldsymbol{0.944\pm 0.004}$ &$\boldsymbol{0.917\pm 0.005}$ &$\boldsymbol{0.970\pm 0.002}$ &$\boldsymbol{0.967\pm 0.002}$ &$\boldsymbol{0.946\pm 0.009}$\\
		\bottomrule
	\end{tabular}
	}
\end{table}

\begin{table}
	\caption{AP of link prediction.}
	\label{AP}
	\centering
	\resizebox{\textwidth}{!}{
	\begin{tabular}{lccccc}
		\toprule
		         & Political blogs  & Kohonen          & CiaoDVD          & WikiVote         & DBLP\\
		\midrule
		PSLSM    &$0.887\pm 0.005$&$0.903\pm 0.002$&$0.943\pm 0.002$& N/A              & N/A \\
		VGAE     &$0.830\pm 0.006$&$0.783\pm 0.007$&$0.846\pm 0.003$&$0.778\pm 0.000$&$0.640\pm 0.000$\\
		Graphite &$0.752\pm 0.011$&$0.839\pm 0.007$&$0.728\pm 0.003$&$0.735\pm 0.002$&$0.549\pm 0.001$\\
		DGLFRM   &$0.887\pm 0.006$&$0.833\pm 0.005$&$0.924\pm 0.002$&$0.901\pm 0.001$&$0.835\pm 0.000$\\
		LGVG     &$0.915\pm 0.005$&$0.859\pm 0.004$&$0.945\pm 0.002$&$0.948\pm 0.000$&$0.909\pm 0.000$\\
		\midrule
		DLSM-IP  &$0.882\pm 0.005$&$0.884\pm 0.005$&$0.965\pm 0.003$&$0.964\pm 0.000$&$0.933\pm 0.000$\\
		DLSM     &$\boldsymbol{0.932\pm 0.009}$ &$\boldsymbol{0.924\pm 0.006}$ &$\boldsymbol{0.967\pm 0.002}$ &$\boldsymbol{0.967\pm 0.002}$ &$\boldsymbol{0.944\pm 0.010}$\\
		\bottomrule
	\end{tabular}
	}
\end{table}

\subsection{Community Detection}


\paragraph{Community detection results} We further conduct community detection on three real world datasets, i.e. Political blogs, Emails and British MP. Specially, we adopt K-means method with ground-truth community numbers for clustering using the learned node embeddings (the latent positions $\z_i$ for DLSM) of all methods. The accuracy is used as the evaluation metric, which is computed as the percentage of correctly clustered samples using most likely mappings between true and predicted clusters. Table~\ref{CD} shows that our proposed model significantly outperforms all other baseline methods. Such results verify that the sparse latent positions learned by DLSM can naturally capture the community structure in a high dimensional space.

\begin{table}
	\caption{Accuracies of community detection.}
	\label{CD}
	\centering
	\begin{tabular}{lccc}
		\toprule
		         & Emails         & Political blogs& British MP\\
		\midrule
		VGAE     &$0.423\pm 0.015$&$0.523\pm 0.002$&$0.593\pm 0.002$\\
		Graphite &$0.672\pm 0.012$&$0.514\pm 0.002$&$0.593\pm 0.002$\\
		DGLFRM   &$0.607\pm 0.010$&$0.505\pm 0.001$&$0.572\pm 0.003$\\
		LGVG     &$0.638\pm 0.013$&$0.739\pm 0.013$&$\boldsymbol{0.823\pm 0.017}$\\
		\midrule
		DLSM     &$\boldsymbol{0.816\pm 0.010}$ &$\boldsymbol{0.879\pm 0.007}$ &$0.811\pm 0.005$\\
		\bottomrule
	\end{tabular}
\end{table}

\begin{figure}[htbp]
	\centering
	\subfloat[DGLFRM]{
		\includegraphics[width=0.33\linewidth]{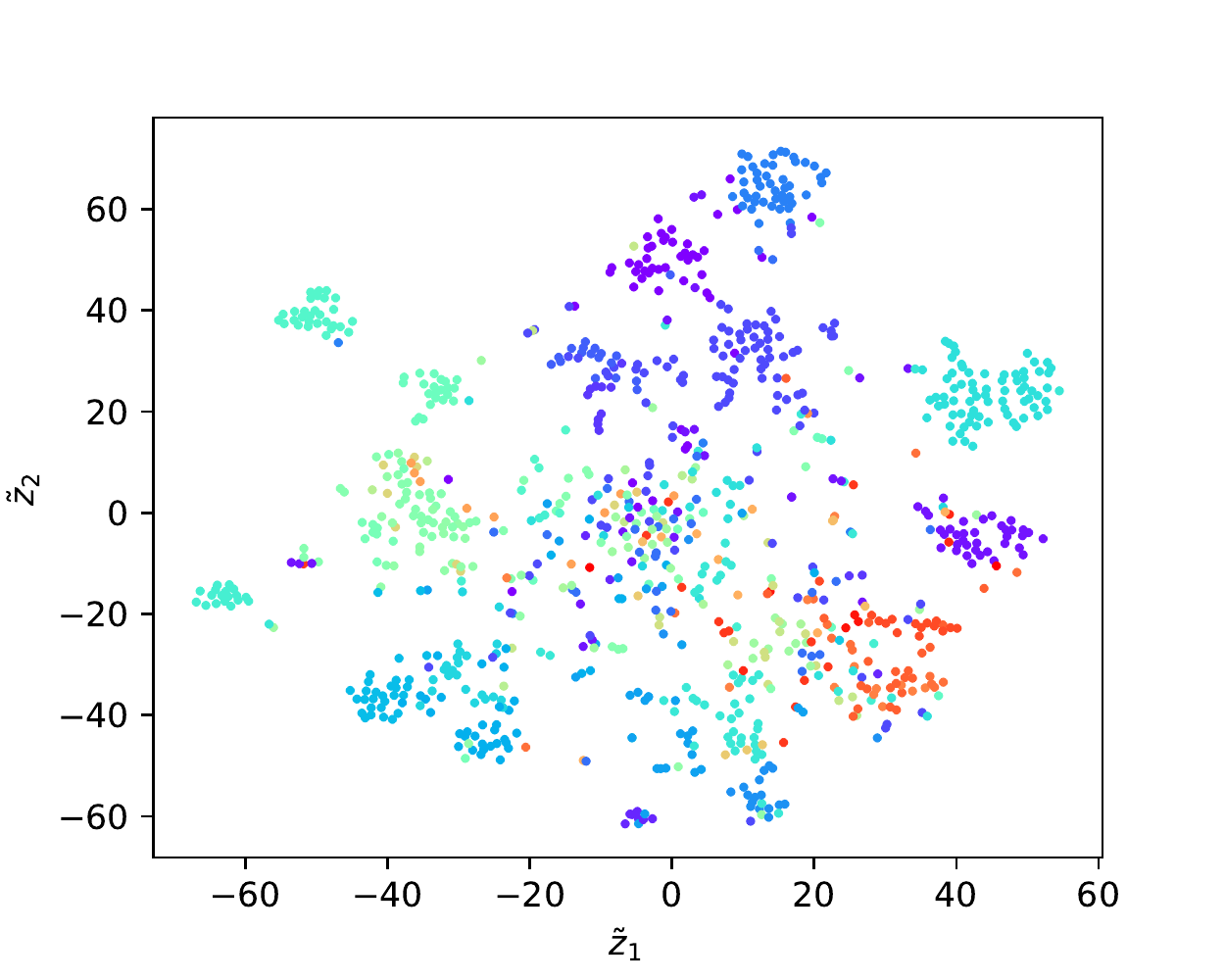}
		}\hspace{-5mm}
	\subfloat[LGVG]{
		\includegraphics[width=0.33\linewidth]{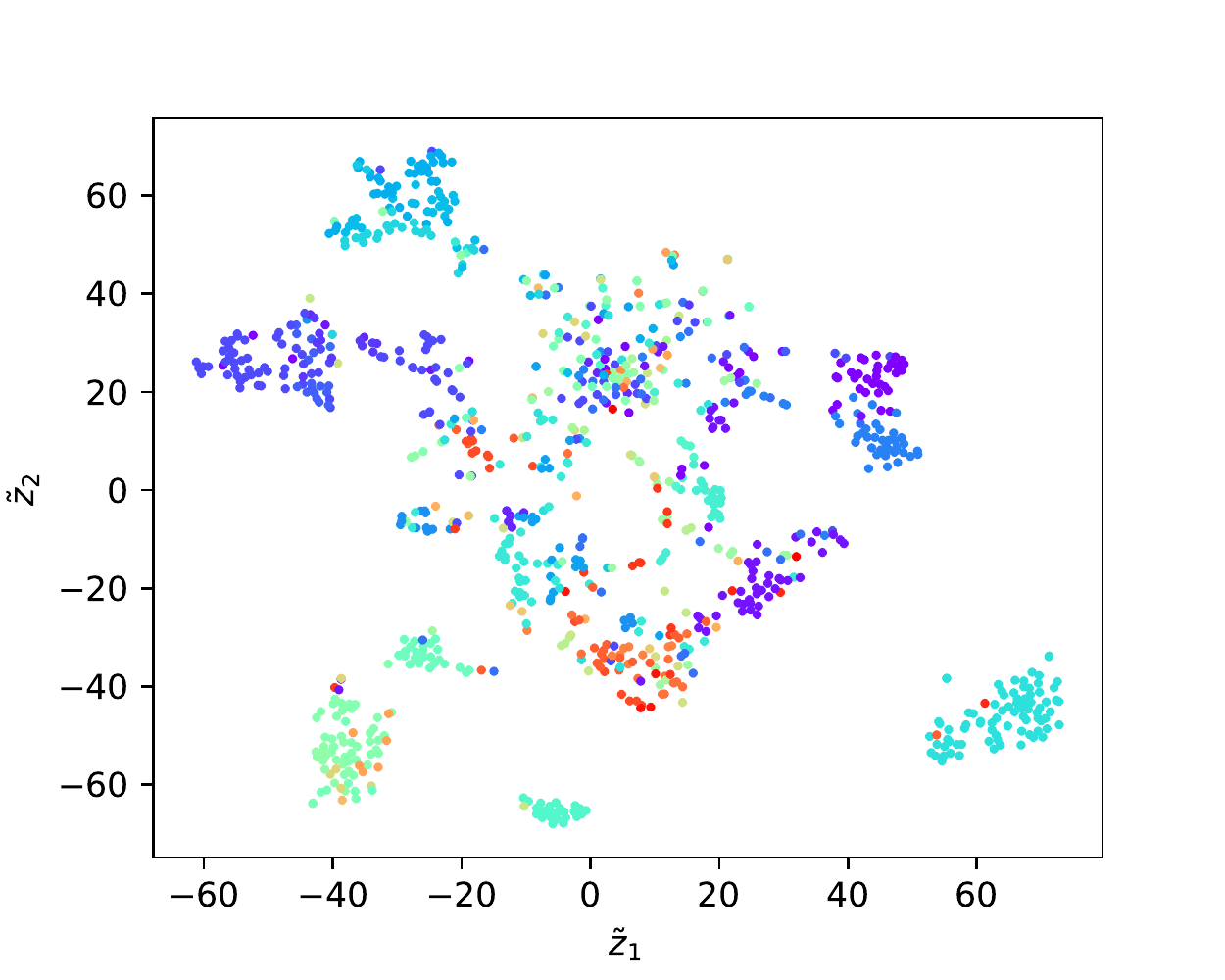}
		}\hspace{-5mm}
	\subfloat[DLSM]{
		\includegraphics[width=0.33\linewidth]{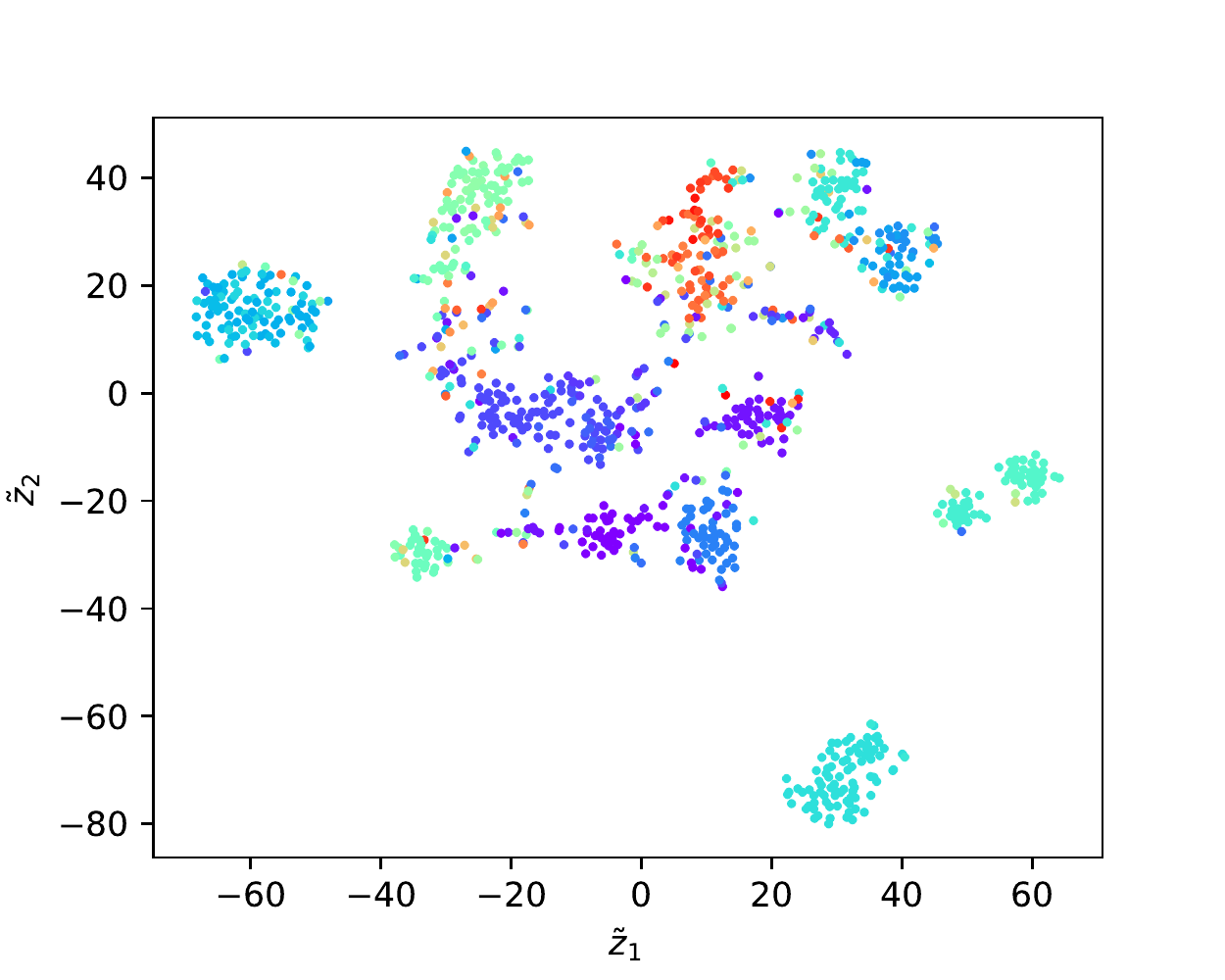}
		}
	\centering
	\caption{Visualizations of the node embeddings learned on the Email network using a 2D t-SNE projection. Colors denote the ground-truth communities.\label{visual}}
\end{figure}

\paragraph{Visualizations of node embeddings}
We leverage a 2D t-SNE projection \cite{van2008visualizing} to visualize the learned latent positions for the Emails dataset. As comparisons, we also illustrate the node representations learned by DGLFRM and LGVG, both of which only consider the community structure but overlook the degree heterogeneity of networks. The transformed latent variables learned by the three models are plotted in Fig.~\ref{visual}. It is clearly seen that our model performs best in fitting such directed networks.

\subsection{Interpretation of node random factors}

\begin{figure}[htbp]
	\centering
	\subfloat[PDD of out-degrees]{
		\includegraphics[width=0.33\linewidth]{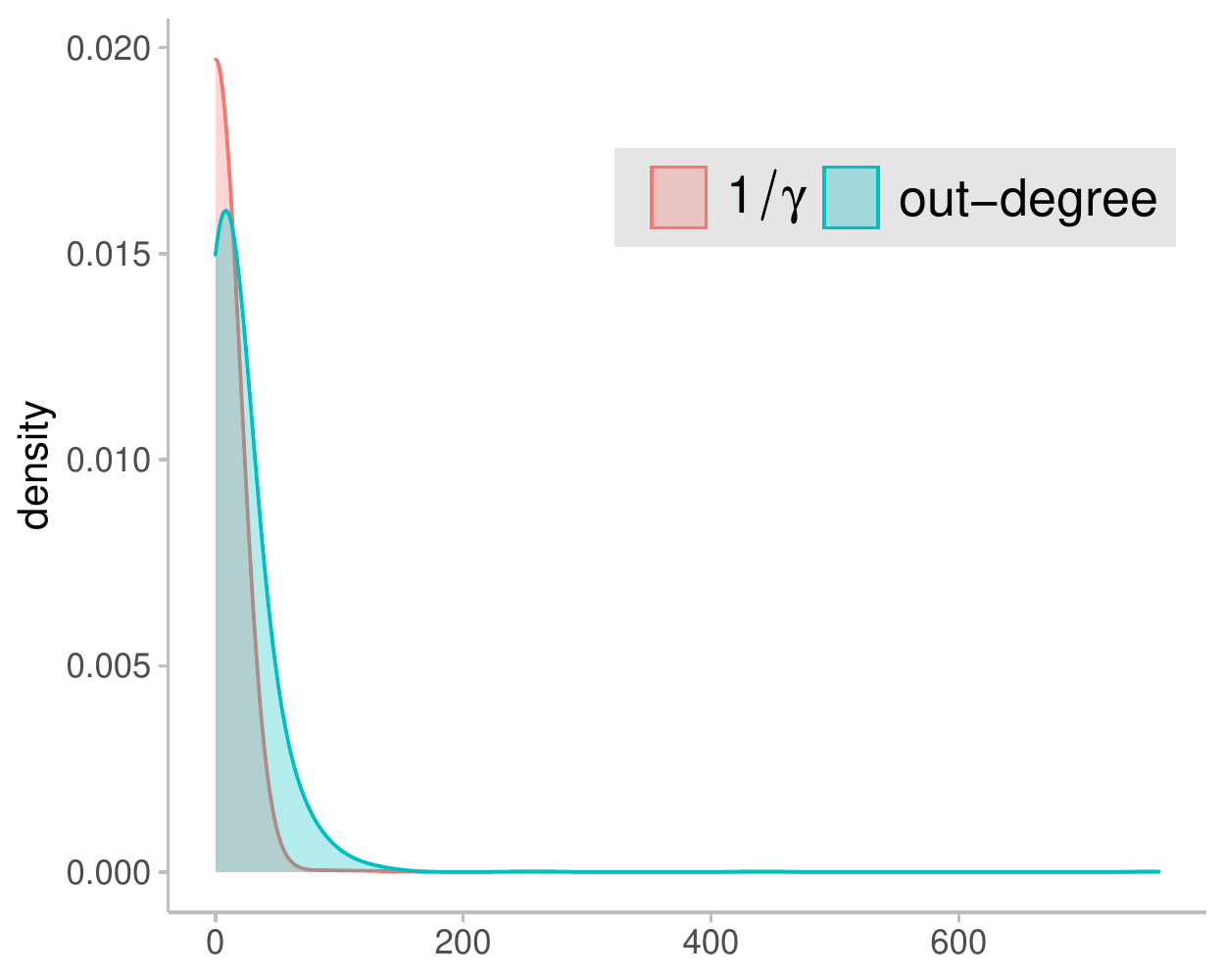}
	}\hspace{-3mm}
	\subfloat[PDD of in-degrees]{
		\includegraphics[width=0.33\linewidth]{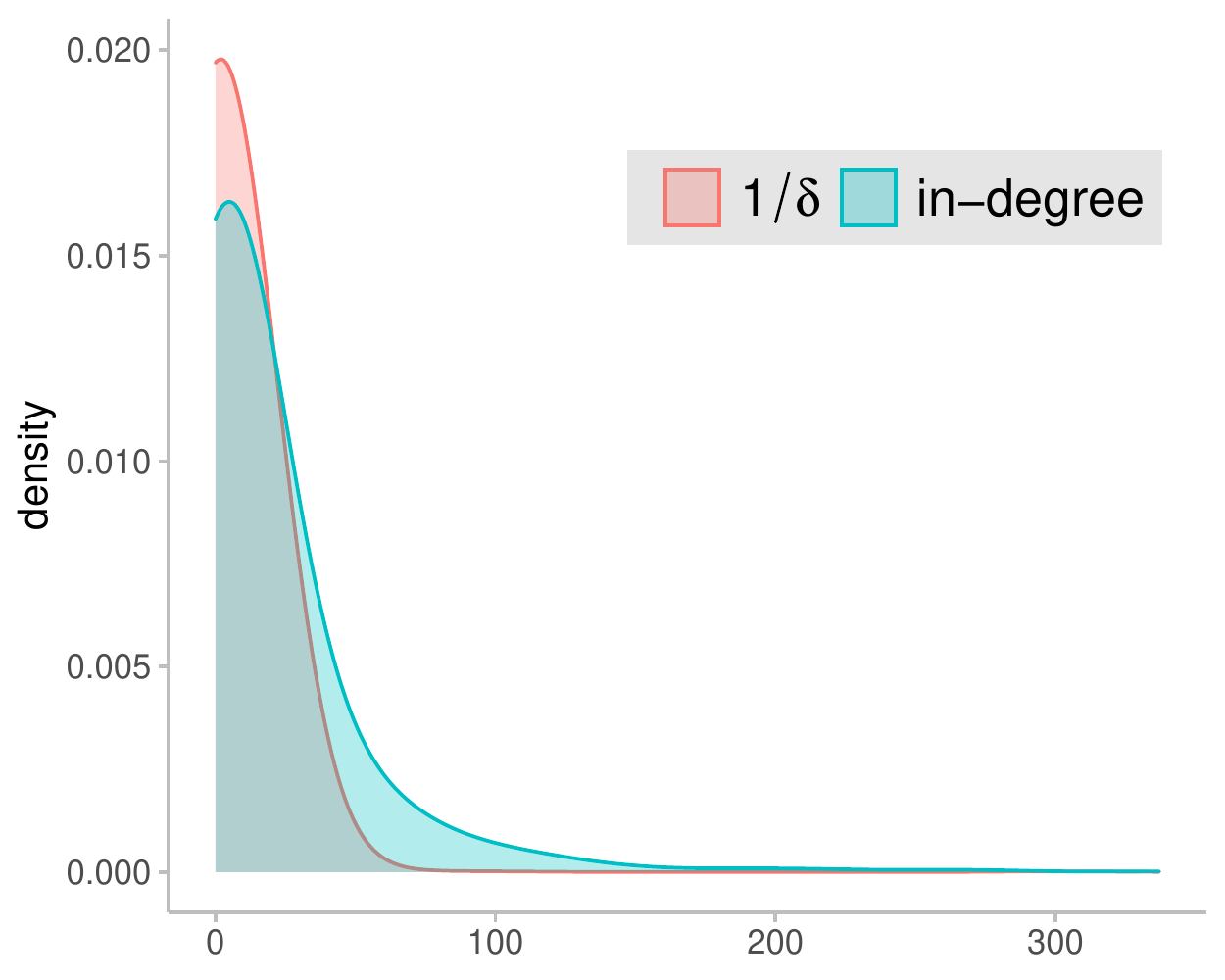}
	}\hspace{-3mm}
	\subfloat[CCD of random factors]{
		\includegraphics[width=0.33\linewidth]{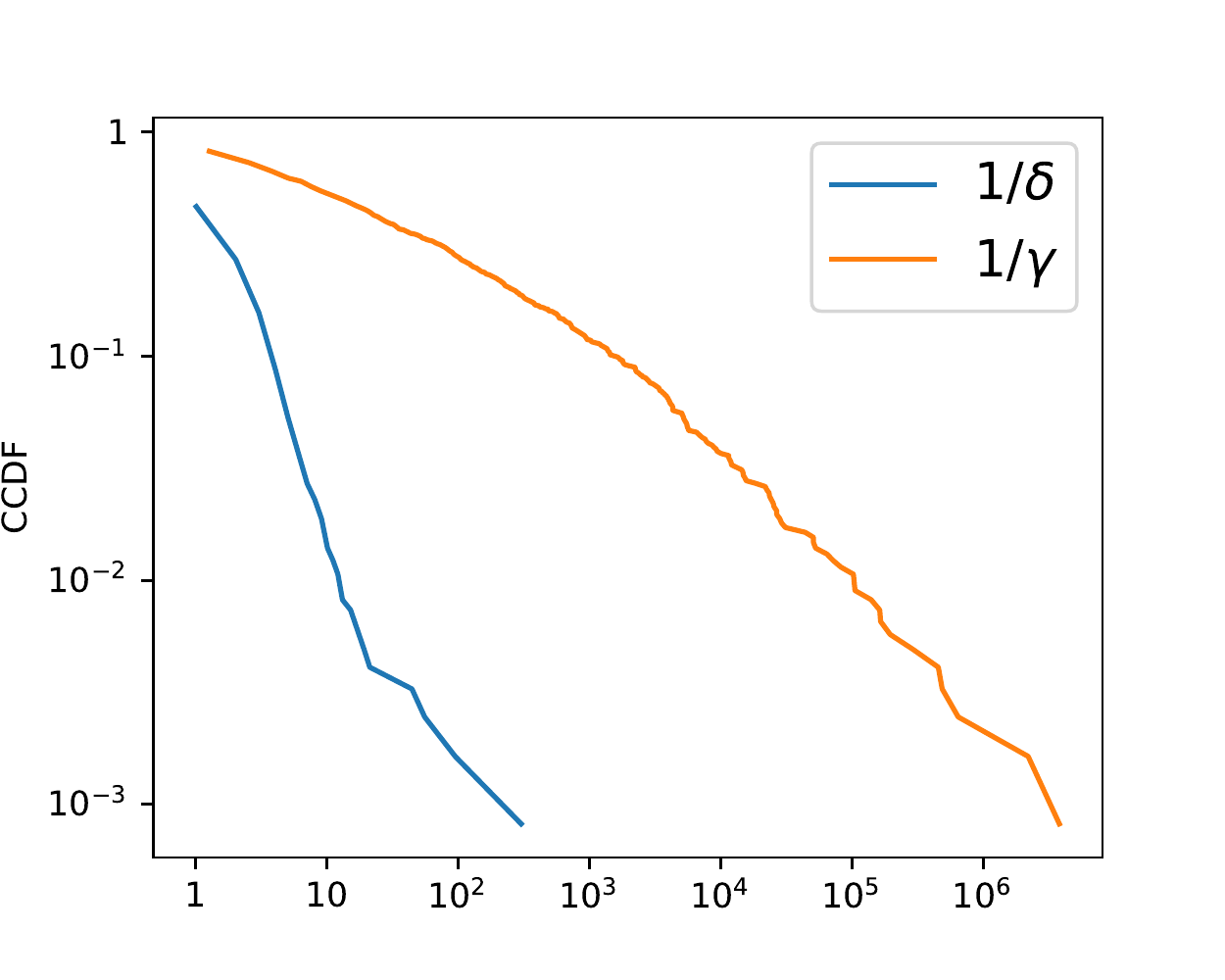}
	}
	\centering
	\caption{Probability density distributions of the degrees and node random factors learned by DLSM.\label{degrees}}
\end{figure}

The pairwise node random factors $\gam_i$ and $\del_i$ are supposed to measure the heterogeneity of out-degrees and in-degrees, respectively, which typically follow the power-law. Fig.~\ref{degrees}(a) and (b) present the probability density distributions (PDD) of the node degrees and reverse random factors learned by DLSM on the political blogs network. It seems that the degree distributions are well fitted by the reverse node random factors, indicating that our DLSM can ideally represent the degree heterogeneity via these latent variables. Furthermore, Fig.~\ref{degrees}(c) illustrates the complementary cumulative distributions (CCD) of the random factors. As can be seen, the logarithm CCD of both $\gam_i$ and $\del_i$ are approximately linear, with different slopes though. This shows that the proposed Dirichlet latent variables are flexible enough to accommodate the power-law distribution of degrees.

\section{Conclusion}

We establish a hierarchical VAE architecture to incorporate the classic LSM into deep learning frameworks. The proposed model, dubbed DLSM, is comprised of a deterministic GNN encoder and a stochastic Bayesian decoder, which is devised for the networks with degree heterogeneity. Series of experiments have shown that the model is valuable in fitting directed networks and achieves the state-of-the-art performance on link prediction and community detection. In addition, the interpretable node embeddings learned by the model can naturally represent both the community structure and degree heterogeneity of complex directed graphs. In the future, we shall extend our model for the more complicated scenes such as weighted or dynamic networks. While the former with multi-valued edges can be simply achieved, the latter demands for a more efficient method to learn the evolutionary topology of graphs.


\bibliographystyle{plain}
\bibliography{ref}

\end{document}